\documentclass[10pt,twocolumn,letterpaper]{article}

\usepackage[pagenumbers]{cvpr} 

\usepackage{graphicx}
\usepackage{amsmath}
\usepackage{amssymb}
\usepackage{booktabs}

\usepackage{subcaption}

\newcommand{\norm}[1]{\left\lVert#1\right\rVert}
\usepackage{multirow}
\usepackage{rotating}
\usepackage{arydshln}
\usepackage{pifont} 
\usepackage{mathabx}

\definecolor{cvprblue}{rgb}{0.21,0.49,0.74}
\usepackage[pagebackref,breaklinks,colorlinks,allcolors=cvprblue]{hyperref}

\def\paperID{18600} 
\def\confName{CVPR}
\def\confYear{2025}

\title{Salient Information Preserving Adversarial Training Improves Clean and Robust Accuracy \thanks{An early version of this paper was first submitted to Winter Conference on Applications of Computer Vision (WACV) 2024 on 6/28/2023.}}

\author{Timothy Redgrave\\
University of Notre Dame\\
{\tt\small tredgrav@nd.edu}
\and
Adam Czajka\\
University of Notre Dame\\
{\tt\small aczajka@nd.edu}
}

\begin{document}

\makeatletter
\let\@oldmaketitle\@maketitle
\renewcommand{\@maketitle}{\@oldmaketitle
  \centering\includegraphics[width=0.9\linewidth]
    {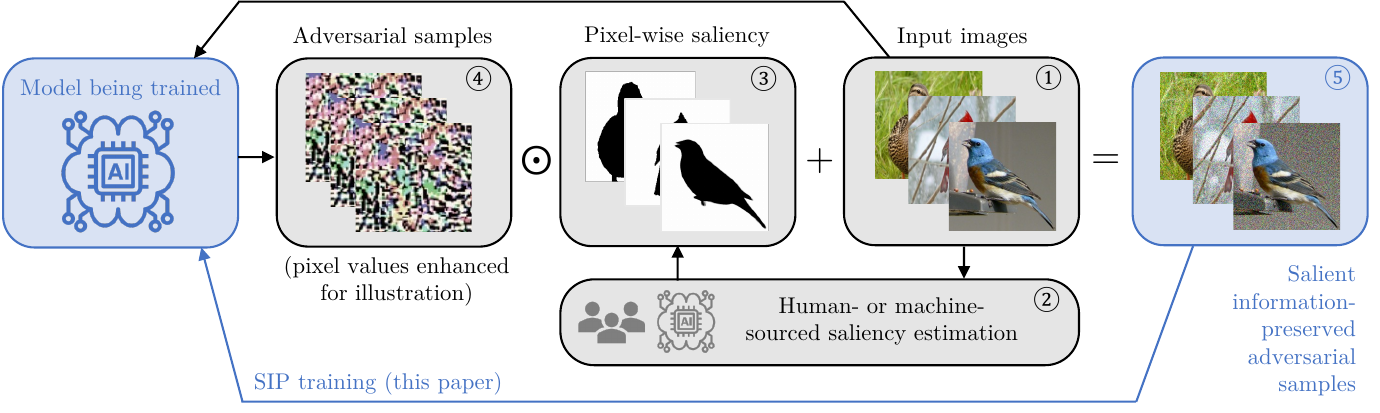}\\
    \captionof{figure}{An illustration of Salient Information Preserving Adversarial Training (SIP-AT). Input images \ding{172} are first given to an annotator \ding{173} (either human or a machine) which generates estimates \ding{174} of which regions of the image should be considered salient. The input images are then fed through the neural network that is to be trained. The standard adversarial samples \ding{175} obtained for this model are combined (via element-wise multiplication) with the salience maps to produce adversarial samples \ding{176} which preserve the salient regions of the original images. These salient information-preserved adversarial samples are then used to train the model.}
    \label{fig:Teaser_Figure}
    \bigskip}
\makeatother

\maketitle
\begin{abstract}
In this work we introduce Salient Information Preserving Adversarial Training (SIP-AT), an intuitive method for relieving the robustness-accuracy trade-off incurred by traditional adversarial training. 
SIP-AT uses salient image regions to guide the adversarial training process in such a way that fragile features deemed meaningful by an annotator remain unperturbed during training, allowing models to learn highly predictive non-robust features without sacrificing overall robustness.
This technique is compatible with both human-based and automatically generated salience estimates, allowing SIP-AT to be used as a part of human-driven model development without forcing SIP-AT to be reliant upon additional human data.
We perform experiments across multiple datasets and architectures and demonstrate that SIP-AT is able to boost the clean accuracy of models while maintaining a high degree of robustness against attacks at multiple epsilon levels.
We complement our central experiments with an observational study measuring the rate at which human subjects successfully identify perturbed images.
This study helps build a more intuitive understanding of adversarial attack strength and demonstrates the heightened importance of low-epsilon robustness.
Our results demonstrate the efficacy of SIP-AT and provide valuable insight into the risks posed by adversarial samples of various strengths.
\end{abstract}    
\section{Introduction}
\label{sec:intro}

Within the past decade, neural networks have come to dominate the field of computer vision due to their unprecedented levels of performance across a variety of tasks~\cite{he2015delving,mckinney2020international,kebria2019deep}. 
However, these systems have repeatedly been proven to be highly susceptible to adversarial attacks -- the intentional modification of inputs in order to cause model failure~\cite{szegedy2014intriguing,goodfellow2014explaining}. 
These vulnerabilities present a serious security risk, especially as neural networks see increased adoption within security critical fields such as medical imaging~\cite{mckinney2020international} or autonomous vehicles~\cite{kebria2019deep}.
In response to this threat, a great deal of effort has been dedicated to study of both adversarial attacks and defenses.
While advances in adversarial attacks has led to increases in attack strength and speed~\cite{carlini2017towards,andriushchenko2020square,croce2020reliable}, adversarial defense methods have lagged behind, with many initially promising defense strategies proving to be vulnerable to more determined or more intelligent adversaries~\cite{he2017adversarial,carlini2017towards,athalye2018obfuscated,Redgrave_2023_CVPR}.

One factor that harms the efficacy of traditional adversarial training is the tendency for adversarially trained models to have lower accuracy on unperturbed images than their traditionally trained counterparts~\cite{su2018robustness,tsipras2018robustness}. 
Prior research into this phenomenon has hypothesized that the trade-off between robustness and accuracy is a consequence of brittle yet predictive features within the training data that are ``wiped out'' by adversarial training~\cite{ilyas2019adversarial}.
Inspired by the idea that meaningful and non-meaningful features can (to an extent) be separated by locality, we propose salient information preserving adversarial training (SIP-AT). 
As shown in Fig. \ref{fig:Teaser_Figure}, SIP-AT prevents the erasure of meaningful features by imposing strict restrictions on the set of feasible image perturbations during the adversarial training process, thereby allowing models to learn predictive features that would otherwise be unavailable.

In this paper we first build a theoretical framework for analyzing SIP-AT and layout a practical approach for implementing SIP-AT.
We then perform experiments over both general and fine-grained image classification tasks and demonstrate that our proposed SIP-AT method improves both clean-image accuracy and maintains or boosts robust accuracy.
We find that these results hold across multiple models and against attacks with varying perturbation budgets (\ie epsilons $\varepsilon$), with SIP-AT providing the greatest improvement against low $\varepsilon$ attacks.
Finally, we compliment these findings with a study evaluating the rate at which humans are able to correctly notice an image has been perturbed; we find that minimally perturbed images are almost always missed by humans, whereas attacks at $\varepsilon=4,8$ are able to be detected at a semi-consistent rate, thereby highlighting the increased importance of low $\varepsilon$ robustness. 
\section{Related Work}
\subsection{Adversarial Training}
Many modern approaches to adversarial training build upon the framework established by Madry \etal~\cite{madry2018towards} which formulates adversarial training as a minimax optimization problem that aims to find the set of parameters that minimize the expected maximum loss across the training data.
In their own paper, Madry \etal~\cite{madry2018towards} perform adversarial training by using Projected Gradient Descent (PGD) to generate adversarial samples at each training step, allowing them to estimate the expected maximum loss. 

Research by Shafahi \etal~\cite{shafahi2019adversarial}, Zhang \etal~\cite{zhang2019you}, and Wong \etal~\cite{wongfast} have proposed methods for drastically reducing the time required for adversarial training.
Experiments performed by Bai \etal~\cite{bai2021transformers} and Zhou \etal~\cite{zhou2022understanding} investigate the relationship between robustness and network architecture in further detail.
Rice \etal~\cite{rice2020overfitting} demonstrate how over-fitting during adversarial training affects test performance more than over-fitting during classic training.
Others such as Xing \etal~\cite{xing2021algorithmic} analyze adversarial training from an algorithmic perspective in order to determine its stability and theoretical upper bound for generalization error.
Further works such as that done by Tramer \etal~\cite{tramer2019adversarial} explore expanding robustness to multiple attack types.

Within the existing adversarial training literature, our work most closely resembles \textbf{P}ixel-reweighted \textbf{A}dve\textbf{R}sarial \textbf{T}raining (PART) \cite{zhang2024improving}.
Proposed by Zhang \etal~\cite{zhang2024improving}, PART attempts to guide a model to focus on key regions during adversarial training by reducing the perturbation budget for pixels that are deemed unimportant according to the class activation mappings (CAMs) generated by the model that is being trained.
While this approach shares a number of superficial similarities to SIP-AT, there are a number of critical differences between the two methods.
The first and most fundamental difference between PART and SIP-AT comes from their radically divergent approaches to using pixel-wise importance maps for guiding adversarial training.
While PART \textit{reduces} the perturbation budget for pixels \textit{outside} of important regions, SIP-AT \textit{zeroes} the perturbation budget for pixels \textit{within} salient regions. 
These two strategies are antithetical in nature and as a result PART and SIP-AT end up constituting two very distinct approaches to adversarial training.

\subsection{Robustness vs. Accuracy}
The trade-off between adversarial robustness and classification accuracy has been explored in a number of ways, including both theoretical and empirical approaches.
Ilyas \etal~\cite{ilyas2019adversarial} analyze the two goals of accuracy and robustness from a geometric perspective of the data, and they prove a theoretical misalignment between the two tasks.
Gilmer \etal~\cite{gilmer2018adversarial} develop a similar theory, hypothesizing that adversarial samples are a consequence of the high-dimensional geometry of the data manifold. Research by Stutz \etal~\cite{stutz2019disentangling} also considers the dimensionality of the data, however, they push back on the theory that robustness and accuracy are conflicting by adopting a distinction between on-manifold and off-manifold adversarial samples.
Zhang \etal~\cite{zhang2019theoretically} explore the robustness-accuracy trade-off by first developing a theoretical upper bound for robust error, and then developing a novel defense strategy which minimizes clean image loss while regularizing for adversarial robustness.
Other groups attempt to reduce the robustness-accuracy trade-off directly.
Zhang \etal~\cite{zhang2020attacks} propose friendly adversarial training (FAT) which uses minimally adversarial samples rather than samples which maximize the loss during training.
\section{Constructing Salient Information Preserving Adversarial Training (SIP-AT)}
Two premises underlay salient information preserving adversarial training. 
First, adversarial training prevents models from learning non-robust features - including those which are useful (\ie legitimately predictive) for clean images - leading to a drop in clean image accuracy for robustly trained models.
Second, in addition to useful/not-useful and robust/non-robust,  features can also be categorized as salient or non-salient based on whether they are important during classification (\eg features in the background of an image would be non-salient whereas features pertaining to the object would be salient). 
Combining these two ideas, SIP-AT restricts the set of feasible perturbations during adversarial training such that salient features remain intact, thereby allowing models to learn both robust features and a subset of influential non-robust features.

\begin{figure*}[t!]
    \centering
    \includegraphics[width=0.95\linewidth]{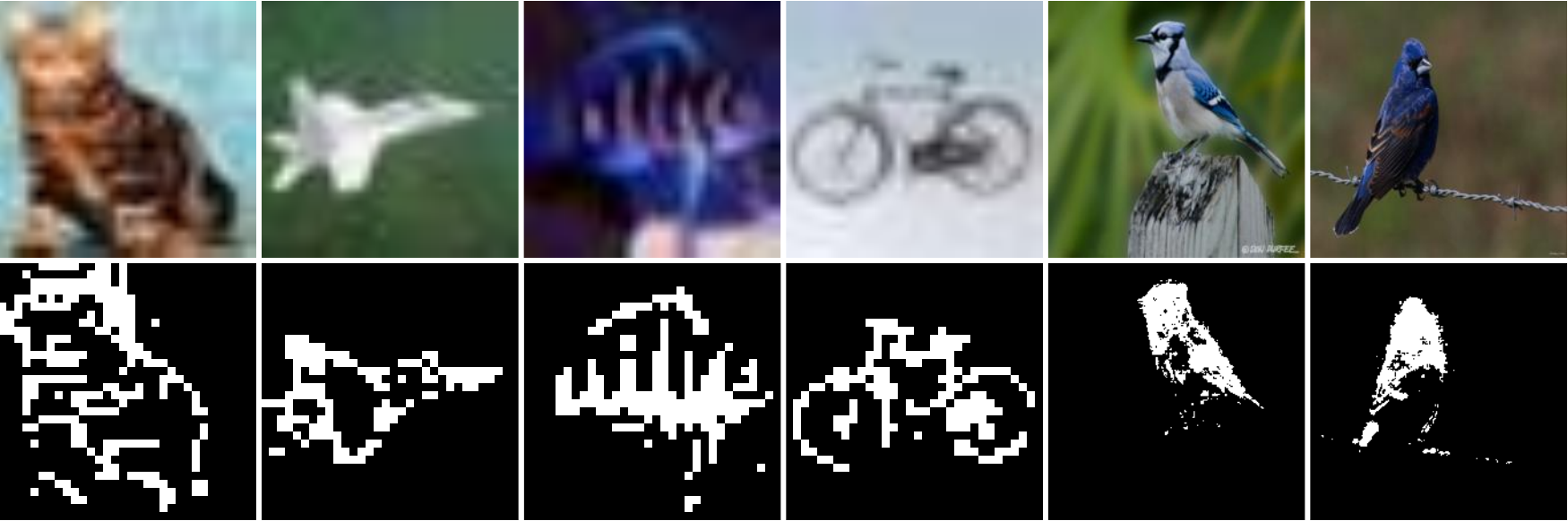}
    \caption{Examples of synthetically generated salience maps following equation \eqref{eq:Top-K_Pixels}. Two samples are shown for each dataset. From left to right: CIFAR-10, CIFAR-100, and CUB-200-2011.}
    \label{fig:synthetic_annotations}
\end{figure*}

\subsection{Theoretical Basis for SIP-AT}
We now build a formal framework for SIP-AT. To this end, we draw from the setting and definitions of Ilyas \etal~\cite{ilyas2019adversarial}.

Consider a binary classification problem where input-label pairs $x,y \in \mathcal{X} \times \{ \pm 1\} $ are drawn from a data distribution $\mathcal{D}$ and the objective is to train a classifier $\mathcal{C}_{\theta}: \mathcal{X} \rightarrow \{ \pm 1\}$ that is parameterized by weights and biases $\theta$ and predicts a label $y$ for a given input $x$. 
Let a \textit{feature} be defined as a function mapping from the input space $\mathcal{X}$ to the real numbers with the set of all features being $\mathcal{F} = \{ f: \mathcal{X} \rightarrow \mathbb{R}\}$.
For convenience features are assumed to be shifted and scaled such that $\mathbb{E}_{(x,y) \sim \mathcal{D}}\left[f(x)\right] = 0$. 

Now let a feature $f$ be categorized as \textit{useful} if it is correlated with the true label expectation (\ie it can be used in a predictive manner).
\begin{equation} \label{eq:useful_feature}
    \mathbb{E}_{(x,y) \sim \mathcal{D}} \left[y\cdot f\left(x\right) \right] > 0
\end{equation}
Further, let $\Delta$ be the set of feasible perturbations that can be applied by a specific adversary. Then, a feature $f$ can be defined to be \textit{robust} against the given adversary if it remains useful under perturbation $\delta\in\Delta$.
\begin{equation} \label{eq:robust_feature}
    \mathbb{E}_{(x,y) \sim \mathcal{D}} \left[ \underset{\delta \in \Delta\left(x\right)}{\text{inf}} y\cdot f\left(x+\delta\right) \right] > 0
\end{equation}
Within our experiments we consider robustness against $L_{\infty}$ bounded adversary with $\Delta \mkern-4mu = \mkern-4mu \{\delta \mid \norm{\delta}_{\infty} \mkern-4mu \varleq \mkern-4mu \varepsilon \}$, but this notion of robustness is applicable to any adversarial threat model. 

We now define salience using a function $S$ (parameterized by a trusted classifier $\mathcal{T}$) that maps from input-feature pairs to $\{0,1\}$ 
\begin{equation} \label{eq:salience}
    S_{\mathcal{T}}: \left(\mathcal{X},\mathcal{F}\right) \rightarrow \{0,1\}
\end{equation}
with $S_{\mathcal{T}}\left(x,f\right) = 0 $ indicating $f$ was unimportant when determining $\mathcal{T}\left(x\right)$ (\ie $f$ is non-salient for the given input) and $S_{\mathcal{T}}\left(x,f\right) = 1$ indicating $f$ influenced the output of $\mathcal{T}\left(x\right)$ (\ie $f$ is salient for the given input). 

Using this definition we can define the set of feasible perturbations that preserve salient features (for a given sample) 
\begin{equation} \label{eq:SIP-AT-Theory}
    \Delta^{\prime} \mkern-4mu \left(x\right) \mkern-4mu = \mkern-4mu \{\delta \mkern-2mu \in \mkern-2mu \Delta \mkern-4mu \left(x\right) \mkern-2mu \mid \mkern-2mu S_{\mathcal{T}} \mkern-2mu \left(x,y,f\right) \mkern-4mu = \mkern-4mu 1 \mkern-2mu \rightarrow \mkern-2mu S_{\mathcal{T}} \mkern-2mu \left(  x \mkern-2mu + \mkern-2mu \delta,y,f\right) \mkern-4mu = \mkern-4mu 1 \}
\end{equation}

Applying this to equation \eqref{eq:robust_feature}, a feature $f$ can be considered to be \textit{salient} or \textit{robust} (inclusive) if it satisfies 
\begin{equation} \label{eq:robust_or_salient}
    \mathbb{E}_{(x,y) \sim \mathcal{D}} \left[ \underset{\delta \in \Delta^{\prime}\left(x\right)}{\text{inf}} y\cdot f\left(x+\delta\right) \right] > 0
\end{equation}

At this point it is evident that the set of robust features  (features satisfying \eqref{eq:robust_feature}) $\mathcal{F}_{robust}$ is a subset of the set of features that are salient or robust (those satisfying equation \eqref{eq:robust_or_salient}) $\mathcal{F}_{sal \vee robust}$ which itself is a subset of all useful features (features satisfying equation \eqref{eq:useful_feature}) $\mathcal{F}_{useful}$.
Consequently, it can be observed that models trained with the restricted set of perturbations $\Delta^{\prime}$ will be able to learn from a larger set of useful features (the set of additional learnable features can be defined explicitly as $\{f \in \mathcal{F}_{sal \vee robust} \setminus \mathcal{F}_{robust}\}$).

At this point we can formally express SIP-AT as a minimax problem.
\begin{equation} \label{eq:SIP-AT_Formalization}
    \underset{\theta}{\textrm{argmin }} \mathbb{E}_{\left(x,y\sim\mathcal{D}\right)} \left[\underset{\delta\in\Delta^{\prime}\left(x\right)}{\text{max}}\mathcal{L}\left(C\left(x\right),y\right)\right]
\end{equation}
With $\mathcal{L}$ being an appropriate loss function such as binary cross-entropy. This formulation ends up being nearly identical to the one used by Madry \etal~\cite{madry2018towards} with the key distinction coming in the form of the restricted set of feasible perturbations $\Delta^{\prime}$.

\subsection{A Practical Approach to SIP-AT}
In practice, it may be difficult or even impossible to explicitly define $S_{\mathcal{T}}$ and measure the contribution of each feature individually (this fact is made plainly apparent if we consider cases where a human performs the role of classifier $\mathcal{T}$ , \eg during data labeling).
As a result, it is beneficial to approximate the set of salient information preserving perturbations $\Delta^{\prime}$ directly.
This can be accomplished by determining which elements within a given input are likely to form the basis of the salient features and then restricting $\Delta$ such that these elements are unperturbed and the salient features remain intact during adversarial training.

For image classification indicating which parts of an image $x$ should be preserved can be done by constructing a salience map $\mathcal{M}\left(x\right)$, where $\mathcal{M}\left(x\right)$ has the same dimensionality as $x$ (we adopt $C,H,W$ for channel, height, and weight) and values of $\{0,1\}$ indicating whether the corresponding element in $x$ should be considered non-salient or salient respectively.
The specific protocol for generating salience maps will be dependent on the classifier used. 
When using a human to classify images and generate salience, salience maps can be constructed by asking the person to annotate an image by drawing to indicate the regions and pixels they found important during classification. 
When using a trained neural network, there are a number of methods that can be used to calculate the contribution of individual pixels to the model's outputs \cite{simonyan2013deep,springenberg2014striving,sundararajan2017axiomatic,shrikumar2017learning}.
For our experiments we generate neural network based salience maps using a rudimentary gradient-based method wherein we select the minimum set of pixels such that they account for over half of the magnitude of the gradient taken with respect to the image.
\begin{equation} \label{eq:Top-K_Pixels}
    \mathcal{M}\left(x\right)_{c,h,w} = 
    \begin{cases}
        1, & \text{if $\nabla_{c,h,w} \in \text{Top-k} \left( \mid \nabla \left( \mathcal{T}\left(x\right) \right) \mid \right)$} \\
        0, & \text{otherwise}
    \end{cases}
\end{equation}
where 
\begin{equation} \label{eq:Minimum_K_for_Top-K_Pixels}
    \text{k} \mkern-3mu = \mkern-3mu \underset{\text{k}}{\text{argmin}} \left[ \sum \text{Top-k} \left( \mid \mkern-3mu \nabla \mkern-3mu \left( \mathcal{T} \mkern-3mu \left(x\right) \right) \mkern-1mu \mid \right) \mkern-2mu \vargeq \mkern-2mu 0.5 \sum \mid \mkern-3mu \nabla \mkern-3mu \left(\mathcal{T} \mkern-3mu \left(x\right)\right) \mkern-1mu \mid \right]
\end{equation}
This method was chosen to highlight that SIP-AT is able to function even when using basic salience estimation strategies.

Regardless of the method used to obtain the salience maps, the process of using them to constrain $\Delta$ is the same.
The set of valid salience preserving perturbations will be 
\begin{equation} \label{eq:SIP-perturbations}
    \Delta^{\prime}\left(x\right) \mkern-3mu=\mkern-3mu \{\delta \in \Delta \mkern-4mu \left(x\right) \mid \mathcal{M}\left(x\right)_{c,h,w} \mkern-3mu = \mkern-3mu 1 \rightarrow \delta_{c,h,w} \mkern-3mu = \mkern-3mu 0\}
\end{equation}
For all $c,h,w\in C,H,W$. 
Fortunately, it is possible to project any valid perturbation $\delta\mkern-2mu\in\mkern-2mu\Delta\left(x\right)$ to the set of salient information preserving perturbations such that $\delta^{\prime}\mkern-2mu\in\mkern-2mu\Delta^{\prime}\left(x\right)$ by applying a mask to the given perturbations $\delta^{\prime} \mkern-4mu = \mkern-4mu \delta \cdot \left( 1 - \mathcal{M}\left(x\right)\right)$.

\section{Experimental Setup}
\subsection{Datasets}
We perform experiments using three datasets.

\noindent \textbf{CIFAR-10}~\cite{krizhevsky2009learning} contains 50,000 training images and 10,000 testing images drawn equally from 10 distinct classes. 

\noindent \textbf{CIFAR-100}~\cite{krizhevsky2009learning} is similar to CIFAR-10 and contains 50,000 training images and 10,000 testing images drawn equally from 100 mutually exclusive classes.

\noindent \textbf{CUB-200-2011}\cite{wah2011caltech,farrell_2022} contains 5,994 training images and 5,794 testing images drawn (approximately equally) from 200 classes, with each class representing a distinct species of birds. CUB-200-2011 also includes a set of human generated segmentation masks which we use as a salience within a subset of our experiments.

By using all three datasets for our experiments we are able to assess the efficacy of SIP-AT for both general image classification (CIFAR-10, CIFAR-100) and fine-grained image classification (CUB-200-2011).
For all three datasets training is performed using a 90-10 training-validation split generated from the training set. 
Likewise, for all three datasets images are normalized to the range $\left[0,1\right]$.
For CIFAR-10 and CIFAR-100, images are kept at their native resolution of (32,32,3).
For CUB-200-2011 images are resized to be (224,224,3).

\subsection{Training Strategies}
In order to assess the efficacy of SIP-AT we train models using six different training strategies:
\begin{itemize}
\item[(a)]  \textbf{No Adversarial Training:} These models are trained in a basic manner (\ie, without any adversarial methods).
\item[(b)] \textbf{Traditional Adversarial Training:} This set of models is trained using the same approach as Madry \etal~\cite{madry2018towards} (\ie, PGD is used to solve the inner minimization problem during training).
\item[(c)] \textbf{TRADES:} This group of models is trained using the TRADES method put forth by Zhang \etal~\cite{zhang2019theoretically}.
TRADES attempts to reduce the robustness-accuracy cost incurred by adversarial training through a trade-off inspired loss regularization. 
\item[(d)] \textbf{Friendly Adversarial Training (FAT):} These models follow the training procedure put forth by \cite{zhang2020attacks}.
FAT is based on the hypothesis that by focusing on the worst case scenario and training using maximally adversarial samples, traditional adversarial training produces overly conservative models.
FAT rectifies this by problem by generating and training against minimally adversarial samples \ie the set of adversarial samples that produce the lowest loss while still leading to misclassification.

\item[(e)] \textbf{Pixel-reweighted AdveRsarial Trainining (PART):} These models use the training strategy of Zhang \etal~\cite{zhang2024improving}. 
During training PART directs models to focus on key regions by reducing the perturbation budget for pixels that are not deemed important to the model according to class activation mapping (CAM) based methods.

\item[(f)] \textbf{Salient Information Preserving Adversarial Training (SIP-AT):} SIP-AT models are trained using the same set-up as traditional adversarial training, but with additional constraints to the set of feasible perturbations as laid out in equation \eqref{eq:SIP-AT_Formalization}.
This results in adversarial perturbations being applied to only non-salient regions during the training process.
When reporting results we use SIP-H and SIP-S to indicate SIP-AT with human salience and SIP-AT with synthetically generated salience respectively.
\end{itemize}

Each training strategy that we choose to use within our experiments was selected on the basis of their ability to help thoroughly assess and analyze SIP-AT. 
The inclusion of models trained without any adversarial elements establish a baseline for clean accuracy. 
Similarly, the models trained with traditional adversarial training serve as a baseline for robust performance.
With these baselines established, TRADES, FAT, and PART can be used as highly informative benchmark methods when attempting to assess the performance of SIP-AT. 
TRADES, FAT, and PART also explicitly aim to address the same problem as SIP-AT (namely alleviating or resolving the trade-off between robustness and accuracy) which makes them even more relevant to the evaluation of SIP-AT.
The use of PART in the assessment of SIP-AT is made even better because not only is PART a recent state-of-the-art method, it also makes use of similar mechanisms to those used by SIP-AT, but with PART employing them a drastically different manner.
Combining these reasons and the methods selected, we argue that our experiments are able to form a strong framework for assessing the performance of SIP-AT.

\subsection{Training Protocols}

\noindent \textbf{CIFAR-10 and CIFAR-100 } We adopt a shared set of training protocols for both CIFAR-10 and CIFAR-100. 
Models are trained using stochastic gradient descent (SGD) for 100 epochs using mini-batches of size 64, an initial training step size of 0.01, and a step decay rate of 0.1 after 60 epochs. 
During training adversaries are given a budget of 10 steps of size $2/255$ and $\varepsilon=8/255$.  
For CIFAR-10 and CIFAR-100 we train both ResNet-18~\cite{he2016deep} and WideResNet-34~\cite{zagoruyko2016wide} models with randomly initialized starting weights. 
Following the observations made by Tsipras \etal~\cite{tsipras2018robustness} and Etmann \etal~\cite{etmann2019connection} that adversarially trained models produce more interpretable salience maps than their non-robust counterparts, we generate the salience maps for SIP-AT following equation \eqref{eq:Top-K_Pixels} using a WideResNet-34 model that was trained via traditional adversarial training.

\noindent \textbf{CUB-200-2011 } For CUB-200-2011 we train ResNet50 ~\cite{he2016deep} and DenseNet121~\cite{huang2017densely} models starting from pretrained ImageNet weights.
SGD is performed for 80 epochs with an initial step size of 0.01 and a step decay rate of 0.75 every ten epochs.
During training, adversaries are given a budget of 32 steps of size $1/255$ and $\varepsilon=8/255$.
For SIP-AT we train models using both human generated salience maps and synthetic salience maps, with the synthetic salience maps being generated from equation \eqref{eq:Top-K_Pixels} using a DenseNet121 model that was trained with traditional adversarial training.

In order to ensure that our findings are consistent across multiple runs, we repeat all experiments five times.

\subsection{Evaluation Protocols}

We test our models using AutoAttack~\cite{croce2020reliable}, a widely used parameter-free approach for evaluating model robustness.
AutoAttack is an ensemble of four diverse adversarial attacks that uses both black-box (Square Attack \cite{andriushchenko2020square}) and white-box attacks (APGD-CE, APGD-DLR, and FAB \cite{croce2020reliable,croce2020reliable,croce2020minimally}).
For our purposes, it is worth noting that because AutoAttack incorporate multiple varied attacks (including Square Attack which does not make use of or require gradients to function) the robustness reported within our results is unlikely to be inflated by obfuscated gradients or other easily circumvented defenses. Consequently, this allows us to place a greater degree of trust that the observed robustness accurately reflects the overall robustness of the training method.

In contrast to many existing works, we assess the robustness of our models across a set of epsilons ($\varepsilon \mkern-4mu=\mkern-4mu 1/255,2/255,4/255,8/255$) rather than at a single fixed epsilon (typically $\varepsilon \mkern-4mu = \mkern-4mu 8/255$).
This allows us to build a more comprehensive understanding of the robustness of each model and provides insight into how model performance degrades as the adversarial budget increases.

It is worth emphasizing that adversaries are free to perturb any/all elements within the images during evaluation (\ie they are not restricted to only non-salient portions of the image).


\subsection{Human Attack-Detection Survey}

\begin{figure}[!ht]
    \begin{center}
    \includegraphics[width=0.95\linewidth]{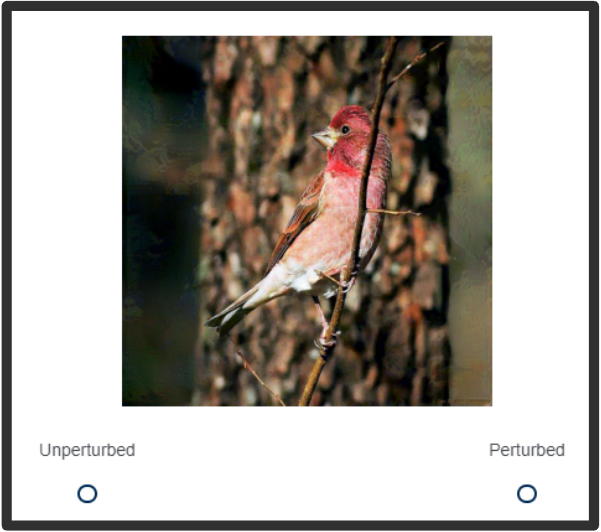}
    \end{center}
    \caption{Example of a question shown to survey participants.}
\label{fig:example_survey_question}
\end{figure}

\begin{table*}
\caption{Top-1 prediction accuracies for CIFAR-10.}
    \centering
    \begin{tabular}{c|l|c|cccc}
    \multicolumn{2}{c}{ } & Clean & \(\varepsilon=1\)& \(\varepsilon=2\) & \(\varepsilon=4\) & \(\varepsilon=8\) \\
    \toprule
    \multirow{6}{8pt}{\begin{turn}{90}ResNet18\end{turn}} 
    & Basic Models 
    &$91.67\mkern-2mu\pm\mkern-2mu0.07$
    &$39.15\mkern-2mu\pm\mkern-2mu0.37$
    &$6.46\mkern-2mu\pm\mkern-2mu0.18$
    &$0.04\mkern-2mu\pm\mkern-2mu0.01$
    &$0.00\mkern-2mu\pm\mkern-2mu0.00$ \\ 
    \cdashline{2-7}
    
    & Madry 
    &$74.39\mkern-2mu\pm\mkern-2mu0.15$
    &$68.33\mkern-2mu\pm\mkern-2mu0.12$
    &$62.14\mkern-2mu\pm\mkern-2mu0.10$
    &$49.59\mkern-2mu\pm\mkern-2mu0.17$
    &$28.40\mkern-2mu\pm\mkern-2mu0.22$ \\
    
    & TRADES
    &$75.52\mkern-2mu\pm\mkern-2mu0.14$
    &$69.01\mkern-2mu\pm\mkern-2mu0.07$
    &$62.09\mkern-2mu\pm\mkern-2mu0.14$
    &$48.47\mkern-2mu\pm\mkern-2mu0.06$
    &$25.14\mkern-2mu\pm\mkern-2mu0.13$ \\

    & FAT 
    &$79.54\mkern-2mu\pm\mkern-2mu0.16$
    &$72.79\mkern-2mu\pm\mkern-2mu0.18$
    &$65.25\mkern-2mu\pm\mkern-2mu0.15$
    &$49.88\mkern-2mu\pm\mkern-2mu0.11$
    &$22.97\mkern-2mu\pm\mkern-2mu0.10$ \\

    & PART
    &$79.99\mkern-2mu\pm\mkern-2mu0.08$
    &$75.34\mkern-2mu\pm\mkern-2mu0.13$
    &$70.33\mkern-2mu\pm\mkern-2mu0.12$
    &$58.80\mkern-2mu\pm\mkern-2mu0.10$
    &$34.89\mkern-2mu\pm\mkern-2mu0.14$ \\
    
    & SIP-S (Ours) 
    &$\textbf{82.34}\mkern-2mu\pm\mkern-2mu\textbf{0.06}$
    &$\textbf{77.69}\mkern-2mu\pm\mkern-2mu\textbf{0.12}$
    &$\textbf{72.40}\mkern-2mu\pm\mkern-2mu\textbf{0.14}$
    &$\textbf{60.15}\mkern-2mu\pm\mkern-2mu\textbf{0.18}$
    &$\textbf{35.10}\mkern-2mu\pm\mkern-2mu\textbf{0.29}$ \\
    \hline
    
    \multirow{6}{8pt}{\begin{turn}{90}WideResNet\end{turn}} 
    & Basic Models 
    &$93.55\mkern-2mu\pm\mkern-2mu0.10$
    &$37.51\mkern-2mu\pm\mkern-2mu0.27$
    &$5.62\mkern-2mu\pm\mkern-2mu0.23$
    &$0.03\mkern-2mu\pm\mkern-2mu0.01$
    &$0.00\mkern-2mu\pm\mkern-2mu0.00$ \\ 
    \cdashline{2-7}
    
    & Madry 
    &$83.85\mkern-2mu\pm\mkern-2mu0.13$
    &$80.07\mkern-2mu\pm\mkern-2mu0.10$
    &$\textbf{75.96}\mkern-2mu\pm\mkern-2mu\textbf{0.18}$
    &$\textbf{66.31}\mkern-2mu\pm\mkern-2mu\textbf{0.31}$
    &$\textbf{45.52}\mkern-2mu\pm\mkern-2mu\textbf{0.62}$ \\
    
    & TRADES 
    &$78.43\mkern-2mu\pm\mkern-2mu0.08$
    &$72.21\mkern-2mu\pm\mkern-2mu0.18$
    &$65.32\mkern-2mu\pm\mkern-2mu0.18$
    &$51.00\mkern-2mu\pm\mkern-2mu0.17$
    &$26.74\mkern-2mu\pm\mkern-2mu0.06$ \\

    & FAT 
    &$83.50\mkern-2mu\pm\mkern-2mu0.14$
    &$76.73\mkern-2mu\pm\mkern-2mu0.15$
    &$69.16\mkern-2mu\pm\mkern-2mu0.09$
    &$52.83\mkern-2mu\pm\mkern-2mu0.12$
    &$24.17\mkern-2mu\pm\mkern-2mu0.09$ \\

    & PART 
    &$78.48\mkern-2mu\pm\mkern-2mu0.13$
    &$73.88\mkern-2mu\pm\mkern-2mu0.06$
    &$68.73\mkern-2mu\pm\mkern-2mu0.13$
    &$57.21\mkern-2mu\pm\mkern-2mu0.31$
    &$33.84\mkern-2mu\pm\mkern-2mu0.35$ \\
    
    & SIP-S (Ours) 
    &$\textbf{86.07}\mkern-2mu\pm\mkern-2mu\textbf{0.12}$
    &$\textbf{81.26}\mkern-2mu\pm\mkern-2mu\textbf{0.13}$
    &$75.73\mkern-2mu\pm\mkern-2mu0.29$
    &$63.04\mkern-2mu\pm\mkern-2mu0.30$
    &$36.50\mkern-2mu\pm\mkern-2mu0.37$ \\
    \bottomrule
    \end{tabular}
\label{tab:CIFAR-10}
\end{table*}

\begin{table*}
\caption{Top-1 prediction accuracies for CIFAR-100.}
    \centering
    \begin{tabular}{c|l|c|cccc}
    \multicolumn{2}{c}{ } & Clean & \(\varepsilon=1\)& \(\varepsilon=2\) & \(\varepsilon=4\) & \(\varepsilon=8\) \\
    \toprule
    \multirow{6}{8pt}{\begin{turn}{90}ResNet18\end{turn}} 
    & Basic Models 
    &$70.90\mkern-2mu\pm\mkern-2mu0.15$
    &$14.42\mkern-2mu\pm\mkern-2mu0.29$
    &$1.82\mkern-2mu\pm\mkern-2mu0.02$
    &$0.04\mkern-2mu\pm\mkern-2mu0.00$
    &$0.00\mkern-2mu\pm\mkern-2mu0.00$ \\ \cdashline{2-7}
    
    & Madry 
    &$53.98\mkern-2mu\pm\mkern-2mu0.12$
    &$47.05\mkern-2mu\pm\mkern-2mu0.14$
    &$40.75\mkern-2mu\pm\mkern-2mu0.14$
    &$29.83\mkern-2mu\pm\mkern-2mu0.12$
    &$14.76\mkern-2mu\pm\mkern-2mu0.14$ \\
    
    & TRADES
    &$54.76\mkern-2mu\pm\mkern-2mu0.28$
    &$47.30\mkern-2mu\pm\mkern-2mu0.24$
    &$41.20\mkern-2mu\pm\mkern-2mu0.18$
    &$\textbf{30.63}\mkern-2mu\pm\mkern-2mu\textbf{0.17}$
    &$\textbf{15.54}\mkern-2mu\pm\mkern-2mu\textbf{0.11}$ \\

    & FAT
    &$58.27\mkern-2mu\pm\mkern-2mu0.14$
    &$49.94\mkern-2mu\pm\mkern-2mu0.28$
    &$42.00\mkern-2mu\pm\mkern-2mu0.38$
    &$28.75\mkern-2mu\pm\mkern-2mu0.65$
    &$11.70\mkern-2mu\pm\mkern-2mu0.67$ \\

    & PART
    &$52.88\mkern-2mu\pm\mkern-2mu0.16$
    &$46.33\mkern-2mu\pm\mkern-2mu0.16$
    &$40.28\mkern-2mu\pm\mkern-2mu0.10$
    &$29.69\mkern-2mu\pm\mkern-2mu0.08$
    &$14.48\mkern-2mu\pm\mkern-2mu0.09$ \\
    
    & SIP-S (Ours) 
    &$\textbf{58.97}\mkern-2mu\pm\mkern-2mu\textbf{0.04}$
    &$\textbf{50.46}\mkern-2mu\pm\mkern-2mu\textbf{0.14}$
    &$\textbf{42.67}\mkern-2mu\pm\mkern-2mu\textbf{0.10}$
    &$29.14\mkern-2mu\pm\mkern-2mu0.16$
    &$11.79\mkern-2mu\pm\mkern-2mu0.04$ \\
    \hline
    
    \multirow{6}{8pt}{\begin{turn}{90}WideResNet\end{turn}} 
    & Basic Models 
    &$74.65\mkern-2mu\pm\mkern-2mu0.09$
    &$14.09\mkern-2mu\pm\mkern-2mu0.19$
    &$1.73\mkern-2mu\pm\mkern-2mu0.05$
    &$0.05\mkern-2mu\pm\mkern-2mu0.01$
    &$0.00\mkern-2mu\pm\mkern-2mu0.00$ \\ 
    \cdashline{2-7}
    
    & Madry
    &$57.39\mkern-2mu\pm\mkern-2mu0.18$
    &$50.67\mkern-2mu\pm\mkern-2mu0.08$
    &$44.17\mkern-2mu\pm\mkern-2mu0.13$
    &$33.22\mkern-2mu\pm\mkern-2mu0.19$
    &$17.23\mkern-2mu\pm\mkern-2mu0.11$ \\
    
    & TRADES
    &$58.13\mkern-2mu\pm\mkern-2mu0.10$
    &$50.79\mkern-2mu\pm\mkern-2mu0.14$
    &$44.73\mkern-2mu\pm\mkern-2mu0.12$
    &$33.82\mkern-2mu\pm\mkern-2mu0.09$
    &$\textbf{18.01}\mkern-2mu\pm\mkern-2mu\textbf{0.13}$ \\

    & FAT
    &$63.02\mkern-2mu\pm\mkern-2mu0.14$
    &$\textbf{55.12}\mkern-2mu\pm\mkern-2mu\textbf{0.13}$
    &$\textbf{47.47}\mkern-2mu\pm\mkern-2mu\textbf{0.10}$
    &$\textbf{34.10}\mkern-2mu\pm\mkern-2mu\textbf{0.12}$
    &$15.49\mkern-2mu\pm\mkern-2mu0.09$ \\

    & PART
    &$51.84\mkern-2mu\pm\mkern-2mu0.09$
    &$45.87\mkern-2mu\pm\mkern-2mu0.08$
    &$40.19\mkern-2mu\pm\mkern-2mu0.17$
    &$30.12\mkern-2mu\pm\mkern-2mu0.12$
    &$14.78\mkern-2mu\pm\mkern-2mu0.14$ \\
    
    & SIP-S (Ours) 
    &$\textbf{63.93}\mkern-2mu\pm\mkern-2mu\textbf{0.11}$
    &$55.03\mkern-2mu\pm\mkern-2mu0.13$
    &$46.74\mkern-2mu\pm\mkern-2mu0.19$
    &$32.11\mkern-2mu\pm\mkern-2mu0.12$
    &$13.00\mkern-2mu\pm\mkern-2mu0.07$ \\

    \bottomrule
    \end{tabular}
\label{tab:CIFAR-100}
\end{table*}

In addition to our core set of experiments, we also perform a study with 330 individuals, recruited via the Prolific platform\footnote{\url{https://www.prolific.co}}, in order to determine the rate at which humans are able to successfully identify that an image has been adversarially perturbed across various \(\varepsilon\) levels.
The goal of this experiment was to associate a better context to \(\varepsilon\) levels, understood through human perceptual capabilities to detect such perturbations.

In our study, participants are first shown an example of an unperturbed image and a highly perturbed image (i.e., \(\varepsilon \mkern-2mu=\mkern-2mu 8/255\)) and informed that they will be shown a series of images, and their task will be to determine whether or not the image they are shown is perturbed.
They are then provided a second example of an unperturbed image but this time paired with a minimally perturbed image (\(\varepsilon \mkern-2mu= \mkern-2mu1/255\)) to demonstrate that it may be difficult to determine whether an image has been perturbed or not.
After receiving these instructions and two examples, participants are then presented with a total of 50 images, and for each image they are asked to determine whether the given image has been perturbed or not.
For any given survey, the set of 50 images shown is made up of 10 subsets, with each subset containing the unperturbed version of an image and perturbed versions of the image at different adversarial allowances, i.e., \(\varepsilon = 1/255, 2/255, 4/255,\textrm{ and } 8/255\).
Survey participants could receive one of two versions of the survey containing images generated by attacking either a ResNet50 model or a DenseNet121 model.

For the survey, we made use of images from the CUB-200-2011 dataset resized to be $224\times 224$.
In order to ensure that classes and perturbation levels were equally represented and shown to participants during the study, we constructed a subset of 1,000 images from the test set by randomly selecting 5 distinct images per class.
We then restricted the pool of potential images a survey taker might see to those within our randomly generated subset.  
\section{Results and Discussion}

\begin{table*}
\caption{Top-1 prediction accuracies for CUB-200-2011.}
    \centering
    \begin{tabular}{c|l|c|cccc}
    \multicolumn{2}{c}{ } & Clean & \(\varepsilon=1\)& \(\varepsilon=2\) & \(\varepsilon=4\) & \(\varepsilon=8\) \\
    \toprule
    \multirow{7}{8pt}{\begin{turn}{90}ResNet50\end{turn}} 
    & Basic Models 
    &$73.72\mkern-2mu\pm\mkern-2mu0.17$
    &$0.15\mkern-2mu\pm\mkern-2mu0.01$
    &$0.00\mkern-2mu\pm\mkern-2mu0.00$
    &$0.00\mkern-2mu\pm\mkern-2mu0.00$
    &$0.00\mkern-2mu\pm\mkern-2mu0.00$ \\
    \cdashline{2-7}
    
    & Madry 
    &$34.63\mkern-2mu\pm\mkern-2mu0.46$
    &$27.79\mkern-2mu\pm\mkern-2mu0.39$
    &$22.19\mkern-2mu\pm\mkern-2mu0.44$
    &$13.37\mkern-2mu\pm\mkern-2mu0.33$
    &$4.59\mkern-2mu\pm\mkern-2mu0.12$ \\
    
    & TRADES 
    &$51.06\mkern-2mu\pm\mkern-2mu0.64$
    &$39.71\mkern-2mu\pm\mkern-2mu0.89$
    &$27.87\mkern-2mu\pm\mkern-2mu1.38$
    &$10.32\mkern-2mu\pm\mkern-2mu1.38$
    &$1.14\mkern-2mu\pm\mkern-2mu0.28$ \\
    
    & FAT 
    &$53.19\mkern-2mu\pm\mkern-2mu0.40$
    &$43.95\mkern-2mu\pm\mkern-2mu0.33$
    &$\textbf{35.26}\mkern-2mu\pm\mkern-2mu\textbf{0.44}$
    &$\textbf{20.89}\mkern-2mu\pm\mkern-2mu\textbf{0.50}$
    &$\textbf{5.60}\mkern-2mu\pm\mkern-2mu\textbf{0.46}$ \\

    & PART
    &$46.68\mkern-2mu\pm\mkern-2mu0.29$
    &$38.00\mkern-2mu\pm\mkern-2mu0.25$
    &$30.07\mkern-2mu\pm\mkern-2mu0.31$
    &$17.54\mkern-2mu\pm\mkern-2mu0.25$
    &$4.19\mkern-2mu\pm\mkern-2mu0.08$ \\
    
    & SIP-H (Ours) 
    &$52.28\mkern-2mu\pm\mkern-2mu0.21$
    &$42.65\mkern-2mu\pm\mkern-2mu0.18$
    &$33.74\mkern-2mu\pm\mkern-2mu0.11$
    &$19.55\mkern-2mu\pm\mkern-2mu0.05$
    &$5.04\mkern-2mu\pm\mkern-2mu0.05$ \\
    
    & SIP-S (Ours) 
    &$\textbf{57.21}\mkern-2mu\pm\mkern-2mu\textbf{0.20}$
    &$\textbf{43.97}\mkern-2mu\pm\mkern-2mu\textbf{0.15}$
    &$31.68\mkern-2mu\pm\mkern-2mu0.15$
    &$13.67\mkern-2mu\pm\mkern-2mu0.13$
    &$1.64\mkern-2mu\pm\mkern-2mu0.08$ \\
    
    \hline
    \multirow{7}{8pt}{\begin{turn}{90}DenseNet121\end{turn}} 
    & Basic Models 
    &$76.88\mkern-2mu\pm\mkern-2mu0.12$
    &$0.05\mkern-2mu\pm\mkern-2mu0.00$
    &$0.00\mkern-2mu\pm\mkern-2mu0.00$
    &$0.01\mkern-2mu\pm\mkern-2mu0.01$
    &$0.00\mkern-2mu\pm\mkern-2mu0.00$ \\
    \cdashline{2-7}
    
    & Madry &$37.44\mkern-2mu\pm\mkern-2mu1.56$
    &$30.76\mkern-2mu\pm\mkern-2mu1.39$
    &$24.99\mkern-2mu\pm\mkern-2mu1.27$
    &$15.97\mkern-2mu\pm\mkern-2mu1.02$
    &$5.68\mkern-2mu\pm\mkern-2mu0.44$ \\
    
    & TRADES 
    &$53.34\mkern-2mu\pm\mkern-2mu0.79$
    &$41.30\mkern-2mu\pm\mkern-2mu2.28$
    &$30.18\mkern-2mu\pm\mkern-2mu3.00$
    &$16.98\mkern-2mu\pm\mkern-2mu2.76$
    &$4.16\mkern-2mu\pm\mkern-2mu1.21$ \\
    
    & FAT 
    &$60.66\mkern-2mu\pm\mkern-2mu0.34$
    &$\textbf{51.35}\mkern-2mu\pm\mkern-2mu\textbf{0.32}$
    &$\textbf{42.46}\mkern-2mu\pm\mkern-2mu\textbf{0.31}$
    &$\textbf{26.65}\mkern-2mu\pm\mkern-2mu\textbf{0.56}$
    &$\textbf{7.83}\mkern-2mu\pm\mkern-2mu\textbf{0.55}$ \\

    & PART 
    &$50.78\mkern-2mu\pm\mkern-2mu0.36$
    &$41.86\mkern-2mu\pm\mkern-2mu0.36$
    &$33.56\mkern-2mu\pm\mkern-2mu0.37$
    &$19.70\mkern-2mu\pm\mkern-2mu0.40$
    &$4.85\mkern-2mu\pm\mkern-2mu0.28$ \\

    & SIP-H (Ours) 
    &$58.29\mkern-2mu\pm\mkern-2mu0.17$
    &$49.19\mkern-2mu\pm\mkern-2mu0.22$
    &$39.99\mkern-2mu\pm\mkern-2mu0.18$
    &$24.56\mkern-2mu\pm\mkern-2mu0.14$
    &$6.70\mkern-2mu\pm\mkern-2mu0.10$ \\
    
    & SIP-S (Ours) 
    &$\textbf{63.32}\mkern-2mu\pm\mkern-2mu\textbf{0.29}$
    &$50.11\mkern-2mu\pm\mkern-2mu0.21$
    &$36.69\mkern-2mu\pm\mkern-2mu0.17$
    &$16.69\mkern-2mu\pm\mkern-2mu0.08$&$1.69\mkern-2mu\pm\mkern-2mu0.03$ \\
    \bottomrule
    \end{tabular}
\label{tab:CUB-200-2011}
\end{table*}

\subsection{Clean and Robust Accuracy of SIP-AT}
Examining the results of our experiments, we note a number of important findings.

First focusing on the accuracies for non-perturbed images, it is clear that \textbf{SIP-AT  significantly boosts clean-image accuracy}.
For every dataset and architecture, the models trained using SIP-AT achieve higher accuracy on unperturbed images than their counterparts trained via other adversarial training methods (cf. the ``Clean'' column in Tables \ref{tab:CUB-200-2011}-\ref{tab:CIFAR-10}). 

Moving to examine the performance of models on perturbed images, we can make two more observations regarding low epsilon ($\varepsilon \mkern-2mu=\mkern-2mu 1/255,2/255$) and high epsilon ($\varepsilon \mkern-2mu=\mkern-2mu 4/255,8/255$) attacks respectively: a) \textbf{SIP-AT models exhibit a high degree of robustness against minimally perturbed images}, and b) while their performance degrades more rapidly at higher epsilons, \textbf{SIP-AT models still maintain comparable or improved levels of robust accuracy at higher epsilons} (cf. the ``$\varepsilon$'' columns in Tables \ref{tab:CUB-200-2011}-\ref{tab:CIFAR-10}).
Across the test results, the accuracy of SIP-AT models approaches that of models trained via traditional adversarial training, with the improvements brought about by SIP-AT decreasing as the strength of the adversarial attacks increase.

Next, by switching to examining the results from an architecture perspective it can be noted that \textbf{SIP-AT is effective across architectures}. While the results indicate that ResNet18 and ResNet50 models receive the greatest benefits from using SIP-AT, there are also significant boosts to the clean and low epsilon accuracies of the WideResNet-34 and DenseNet121 models trained using SIP-AT.

\begin{figure}[!ht]
    \begin{center}
    \includegraphics[width=0.95\linewidth]{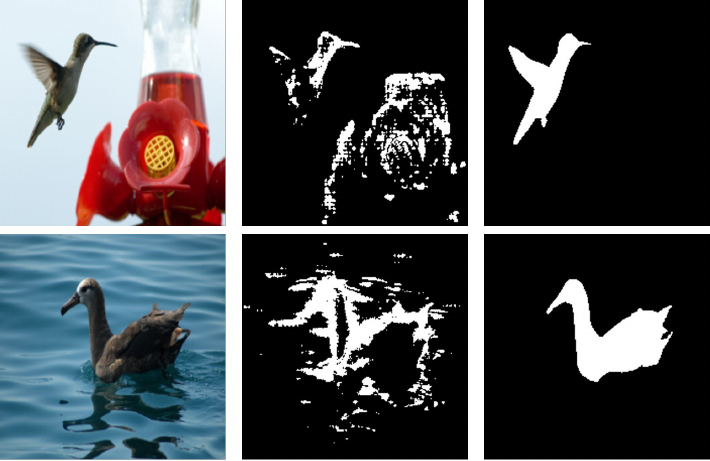}
    \end{center}
    \caption{Examples comparing synthetic salience maps (middle column) against human generated salience maps (right column) for CUB-200-2011. For each image, the synthetic salience maps include regions that - while potentially useful or predictive - are external to the bird itself (\eg, while having a hummingbird feeder within an image may be correlated with the image having the label ``hummingbird'', the presence of the hummingbird feeder has no influence on the type of bird in the picture). In contrast, the human annotations are drawn exclusively on the bird.}
 
\label{fig:Human_vs_Synthetic_Comparison}
\end{figure}

Finally, focusing specifically on the results of the CUB-200-2011 tests, we can observe that \textbf{the type of salience used during training has a significant effect on the behavior of SIP-AT}.
SIP-AT models trained with the use of the synthetically generated salience maps exhibited higher clean accuracy but lower robust accuracy than their counterparts trained with human generated salience maps.
Figure \ref{fig:Human_vs_Synthetic_Comparison} shows examples of how these salience maps differ and how these differences may result in the observed differences in performance.

\subsection{Human Detection of Adversarial Attacks}
\begin{figure}[!ht]
    \begin{center}
    \includegraphics[width=1\linewidth]{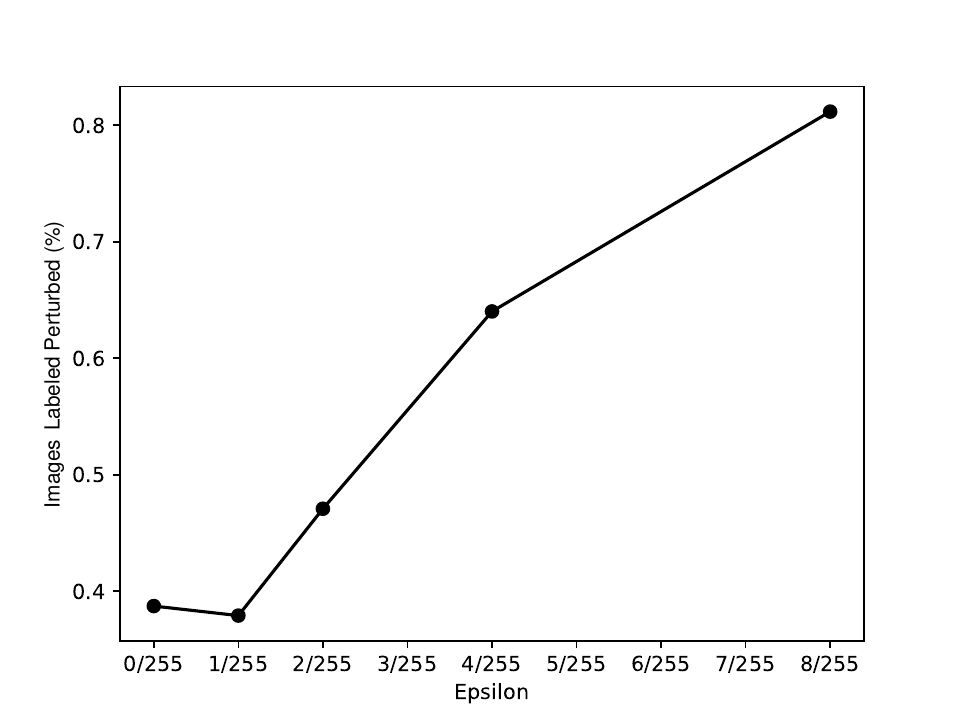}
    \end{center}
    \caption{The rate at which humans indicate an image has been perturbed for varying degrees of attack strength $\varepsilon$. Here $\varepsilon = 0$ indicates clean or unperturbed images.}

\label{fig:human_adv_detection_rate}
\end{figure}

The detection rates achieved by human subjects for various perturbation strengths ($\varepsilon$ levels) are shown in Fig. \ref{fig:human_adv_detection_rate}.
These human trials also yield a number of interesting results. 

First, and rather obvious, \textbf{humans perform poorly for small perturbations.}
These findings are inline with an intuitive analysis of adversarial samples, but they provide support indicating just how dangerous low-epsilon attacks are. 

Second, \textbf{highly perturbed images can be successfully identified.}
In contrast to minimally perturbed images, highly perturbed images (i.e., \(\varepsilon \vargeq 4/255\)) are successfully identified as being perturbed a majority of the time, with observer success rates reaching over 80\% when \(\varepsilon =8/255\).
These results suggest that having the set of feasible perturbations \(\Delta\) include perturbations as large as \(\varepsilon =8/255\) may result in the generation of adversarial examples that are too easily noticed to be considered truly successful adversarial attacks.

Consequently, \textbf{respondents' poor performance on low-epsilon adversarial samples and drastically improved performance on high-epsilon adversarial examples indicates that low-epsilon perturbations pose a greater risk than high-perturbation epsilons.}
This observation is especially important for systems wherein models are deployed with human operators who could spot such anomalies.
This idea can be more firmly understood when considering model failure in a general sense: explainable mistakes can be recovered and learned from, unexplained mistakes cannot; while model failures at $\varepsilon = 8/255$ may remain incorrect and far from ideal, their cause can in the vast majority of cases be identified even by untrained human operators.
\section{Conclusions}
In this paper, we introduced Salient Information Preserving Adversarial Training (SIP-AT).
SIP-AT utilizes information about the location of salient features within images to preserve a collection of non-robust useful features during adversarial training.
We formulate SIP-AT such that it is agnostic to the method used to calculate salience, and we demonstrate the efficacy of SIP-AT for both human and automatically generated estimates of salience, with our SIP-AT models achieving greater clean accuracy and comparable robust accuracy to other adversarial training methods with similar implementation costs.
We hope this new method drives further research in salience informed adversarial training and can be a useful addition to adversarial training toolkits and the study of adversarial robustness.

{
    \small
    \bibliographystyle{ieeenat_fullname}
    \bibliography{main}
}


\end{document}